\documentclass[letterpaper, 10 pt, conference]{ieeeconf}  

\IEEEoverridecommandlockouts                              

\usepackage{times}

\usepackage[numbers]{natbib}
\usepackage{multicol}
\usepackage{color}
\usepackage[utf8]{inputenc}
\usepackage{amsmath,amssymb}
\usepackage{graphicx}
\usepackage{import}
\usepackage{dsfont}
\usepackage{todonotes}
\usepackage{amsfonts}
\usepackage{algorithm}
\usepackage{mathtools}
\usepackage{acronym}
\usepackage{comment}
\usepackage{algpseudocode} 
\usepackage{tabularx}
\usepackage[most]{tcolorbox}
\usepackage[bookmarks=true]{hyperref}

\newif\ifuseboldmathops
\newif\ifuseittextabbrevs
\useboldmathopstrue   

\ifuseittextabbrevs

	\newcommand{\ie}{{\it i.e.}}

\else

	\newcommand{\ie}{i.e.}

\fi

\ifuseboldmathops

\else

\fi

\ifuseboldmathops

\else

\fi

\ifuseboldmathops

\else

\fi

\ifuseboldmathops

\else

\fi

\acrodef{mdp}[MDP]{Markov Decision Process}
\acrodef{sc-rmax}[SC-Rmax]{State Clustering Rmax}
\acrodef{rl}[RL]{Reinforcement Learning}
\acrodef{em}[EM]{expectation-maximization}
\acrodef{ram}[RAM]{relocation action model}
\acrodef{pacmdp}[PAC-MDP]{Probably Approximately Correct MDP}
\acrodef{pac}[PAC]{Probably Approximately Correct}
\acrodef{mle}[MLE]{Maximum Likelihood Estimator}
\acrodef{dof}[DOF]{degree of freedom}
\acrodefplural{dof}[DOFs]{degrees of freedom}
\acrodef{fem}[FEM]{Finite Elements}

\acrodef{gmm}[GMM]{Gaussian Mixture Model}

\renewcommand{\v}[1]{\mathbf{#1}}

\newcommand{\gv}[1]{\ensuremath{\mbox{\boldmath$ #1 $}}} 

\newcommand{\q}{\v{q}}
\newcommand{\qdot}{\dot{\v{q}}}
\newcommand{\qddot}{\ddot{\v{q}}}

\let\oldcomment\Comment
\renewcommand{\Comment}[1]{{\color{blue}\oldcomment{#1}}}

\newcommand{\figref}[1]{Fig. \ref{#1}}

\title{\bf A Validated Physical Model For Real-Time Simulation of Soft Robotic Snakes}

\author{
\begin{tabular*}{1.0\textwidth}{ccccccc}
{Renato Gasoto\textsuperscript{*1,2}} &
{Miles Macklin\textsuperscript{*2,3}} &
{Xuan Liu\textsuperscript{1}} &
{Yinan Sun\textsuperscript{1}} &
{Kenny Erleben\textsuperscript{3}} &
{Cagdas Onal\textsuperscript{1}} &
{Jie Fu\textsuperscript{1}}
\end{tabular*}
\thanks{
\textsuperscript{1} Renato Gasoto, Xuan Liu, and Yinan Sun are PhD students under the supervision of Jie Fu and Cagdas Onal in the Robotics Engineering program at Worcester Polytechnic Institute. \textsuperscript{2}Renato Gasoto and Miles Macklin are affiliated with NVIDIA. \textsuperscript{3}Miles Macklin is a PhD student under the supervision of Kenny Erleben at the University of Copenhagen.  This work was supported in part by the Brazilian National Council for Scientific and Technological Development (CNPq) under grant 208053/2014-0, by the National Science Foundation under grant \#1728412, and by NVIDIA.\textsuperscript{*}Authors have equal contribution in paper.}
}

\begin{document}


%

\maketitle
\begin{abstract}
In this work we present a framework that is capable of accurately representing soft robotic actuators in a multiphysics environment in real-time. We propose a constraint-based dynamics model of a 1-dimensional pneumatic soft actuator that accounts for internal pressure forces, as well as the effect of actuator latency and damping under inflation and deflation and demonstrate its accuracy a full soft robotic snake with the composition of multiple 1D actuators. We verify our model's accuracy in static deformation and dynamic locomotion open-loop control experiments. To achieve real-time performance we leverage the parallel computation power of GPUs to allow interactive control and feedback.

\end{abstract}




\section{Introduction}
\label{sec:introduction}


Soft robotic systems can provide many advantages over rigid robots. For example, the compliance of a soft manipulator allows safe interaction with fragile objects, such as uncooked eggs \cite{ilievski2011soft,trivedi2008soft}. A soft robotic snake can navigate through narrow passages to perform tasks in regions unreachable with rigid robots \cite{luo2014theoretical}.
 On the other hand, soft robots are harder to model and simulate than rigid robots, due to their infinite \acp{dof}. Thus, one often has to compromise between simulation accuracy and simulation time. In recent work \cite{coevoet2017software,rodriguez.coevoet.ea17}, the authors developed a framework of deformable online simulation that can achieve real-time simulation using relatively coarse meshes. Further, the authors proposed a model order reduction method by running expensive offline simulations and applying machine-learning techniques on the generated dataset \cite{goury:hal-01834483}.

However, the compromise between accuracy and simulation time is often hard to make. An accurate model is important to bridge the gap between simulation and reality for control design. On the other hand, online trajectory planning and model predictive control require the prediction of future states with low-latency. In addition, even learning-based controllers, such as those trained with \ac{rl} \cite{trpo,gae,ppo}, typically require thousands of samples to converge, making efficient simulation crucial. Approximate models can enable faster simulations; Furthermore, even in rigid-body simulators, the modeling errors, inaccuracies, and lack of actuator dynamics on the simulator create a gap to the reality that often needs to be bridged before applying controllers learned in simulation to the real robot \cite{neunert2017off}.  

In this work, we aim to develop a real-time, high-fidelity simulation for soft robotic systems. We present a compliant constraint dynamics model for a soft snake robot \cite{snake_1, snake_2, snake_3}. Our model is able to accurately represent the deformations of the snake links while being efficient enough for real-time simulations. To achieve low-latency simulation, we use a GPU-based physics simulator that leverages large-scale parallel iterative solvers to efficiently solve large systems \cite{nvidia2014cusparse}. The resulting simulation is validated against a real robotic snake to verify that the deformation model is accurate and that the dynamics of the simulation match that of the real system. In addition, we measure and validate the actuator and inflation latency to ensure accurate control response. In summary, our main contributions are:


\begin{itemize}
\item A dynamics and actuation framework for 1D pneumatic soft actuators that accurately represents a large range of deformations;
\item A model for a modular soft robotic snake that accurately represents its dynamics;
\item Model validation in static deformation and dynamic locomotion tests;
\item A simulator framework suitable for performing real-time control of soft robots.
\end{itemize}


\section{Related Work}
The field of soft robots includes a large and varied range of designs that incorporate compliant materials and actuators in various forms. Soft robots may have a few flexible regions in them and may be driven by tendons \cite{manti2015bioinspired}, while completely soft pneumatic actuators can be driven by exerting a range of pressures within deformable bodies. The use of pressure leads to many intricate designs that exploit the material geometry to achieve the desired actuation \cite{ilievski2011soft,878151}. 
To allow a large range of pressures and material deformations, hybrid materials are often used on the pressure chambers, such as inextensible layers and fiber reinforcements \cite{7110394}. Our modular soft snake robots use hybrid materials \citep{snake_2,snake_3} to achieve highly efficient 2D terrain locomotion. 


The diversity of soft materials and difficulty in accurate modeling of soft robots lead to an increasing need for building a realistic simulation for deformable robots. Duriez et al. \citep{duriez.13}  presented a framework for simulating soft robots using a quasi-static approach based on \ac{fem}. In this work, we also use \ac{fem} with tetrahedral elements as a fundamental building-block. In addition, we combine these elements in a multiphysics system with spring networks, frictional contact, attachment constraints for soft/hybrid materials and articulated rigid bodies. We perform implicit time-integration to simulate dynamic trajectories. Recent work by Pozzi et al. \cite{pozziefficient} used a rigid-body model, fitted to an offline \ac{fem} simulation augmented with stiff springs to achieve real-time updates. In this work, we simulate the finite-element models and spring networks directly, using the large-scale parallelism of graphics processing units (GPUs) to achieve online update rates, therefore not requiring an expensive offline simulation and data fitting stage.

Tan et al. \citep{Sim-to-real} propose several techniques for bridging the reality gap in simulations. Among them there is a latency introduced for the delay between when a command is sent and when it's executed. Since soft actuators have an over-damped response, our work introduces system dynamic model response to the command on top of the latency.

\section{Soft Robotic Snake}
\label{sec:soft-snake}

\begin{figure}
    \centering
    \begin{tcolorbox}[size = minimal,sharp corners ,blanker, halign=center, halign lower=center]
        \centering
        \vspace{0.15cm}
        \begin{tcolorbox}[size = minimal, sidebyside, righthand width = .5\linewidth,sharp corners ,blanker, halign=center, halign lower=center]
            \centering
            \includegraphics[width=.8\linewidth,angle=-90]{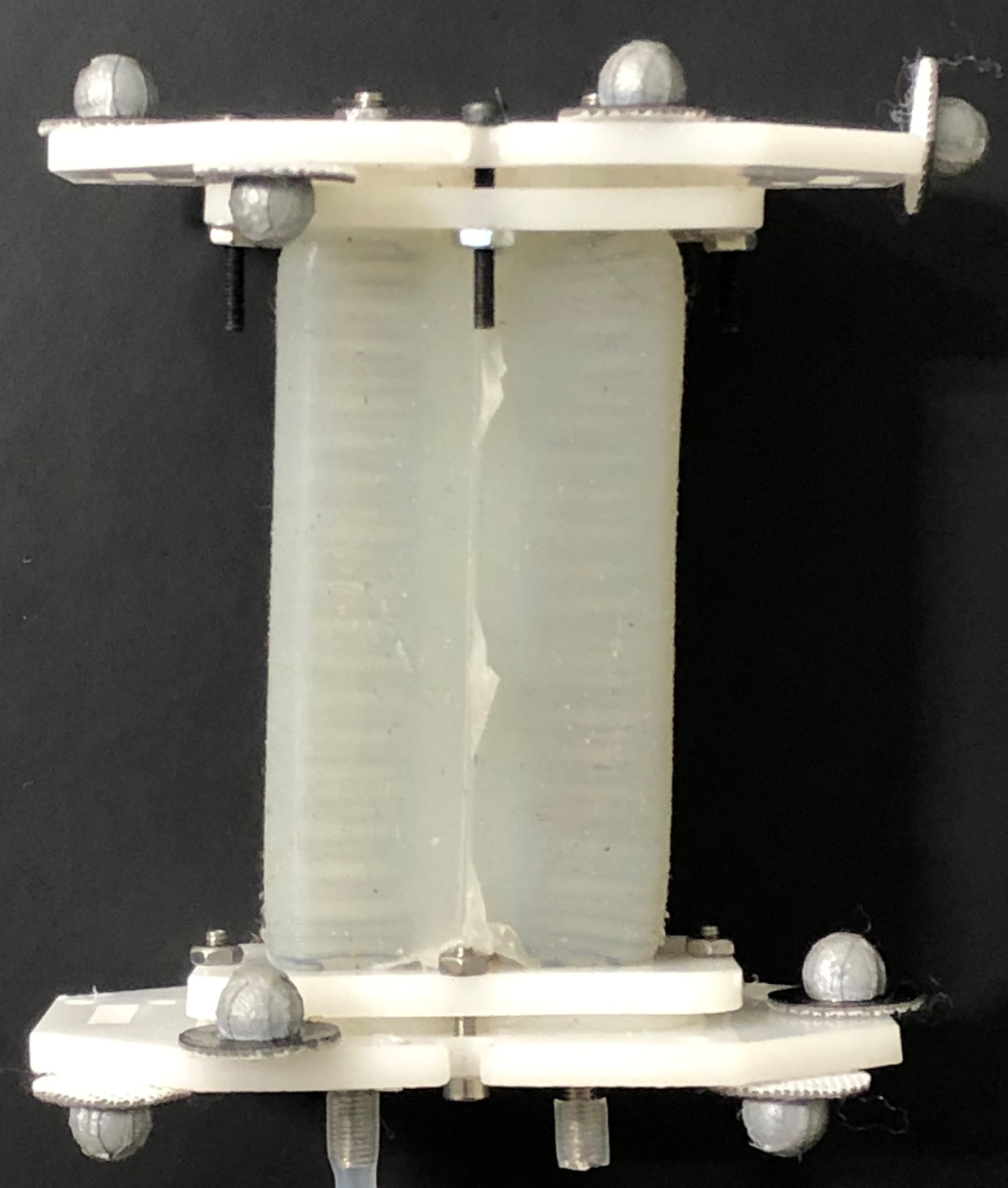}\\   
            \small{(a)}
            \tcblower
            \centering
            \includegraphics[width=.6\linewidth,angle=-90]{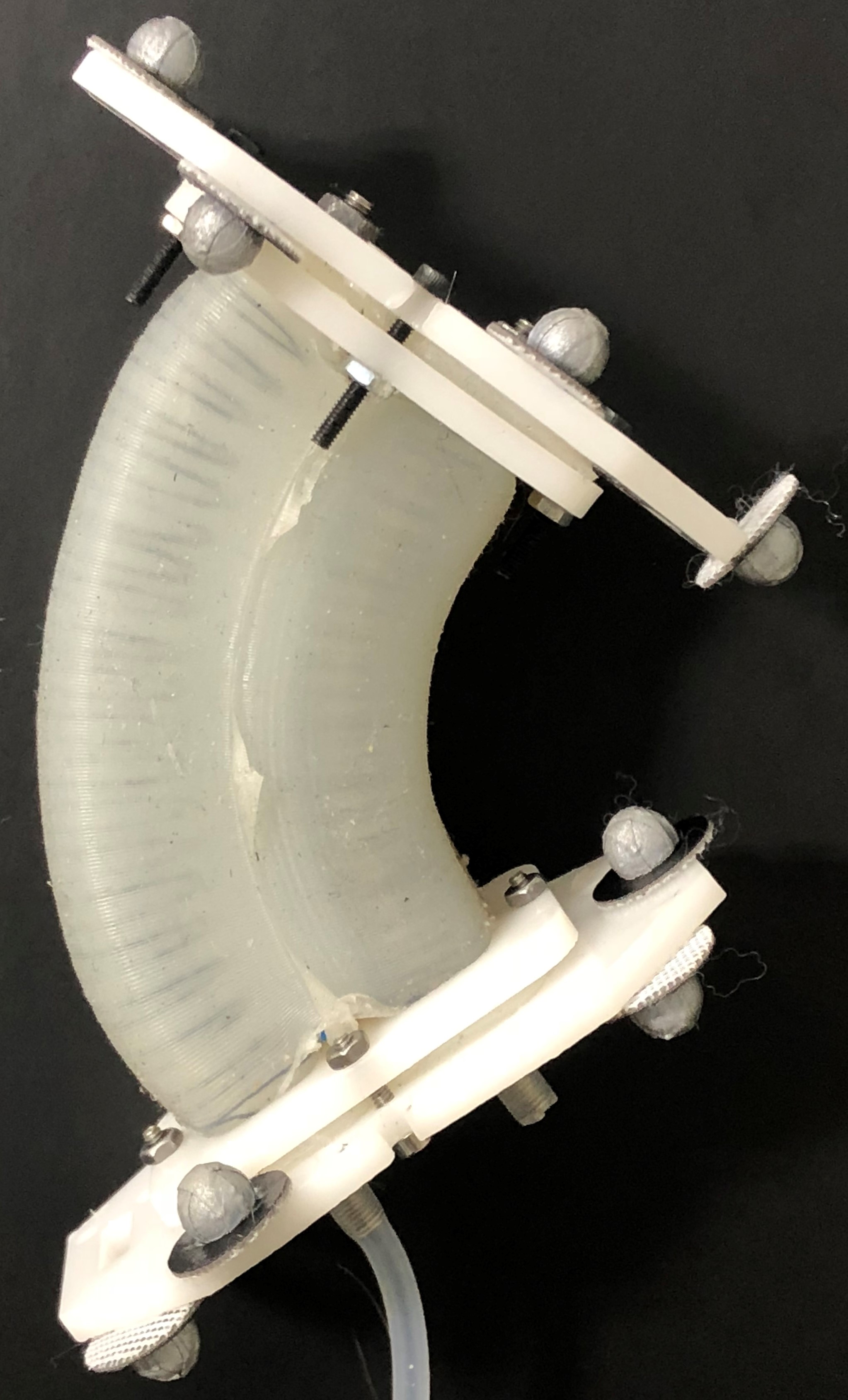}\\
            \small{(b)}
        \end{tcolorbox}
        \tcblower
        \centering
        \begin{tcolorbox}[size = minimal, sidebyside, righthand width = .41\linewidth,sharp corners ,blanker, halign=center, halign lower=center]
            \centering
            \includegraphics[width=1.0\columnwidth]{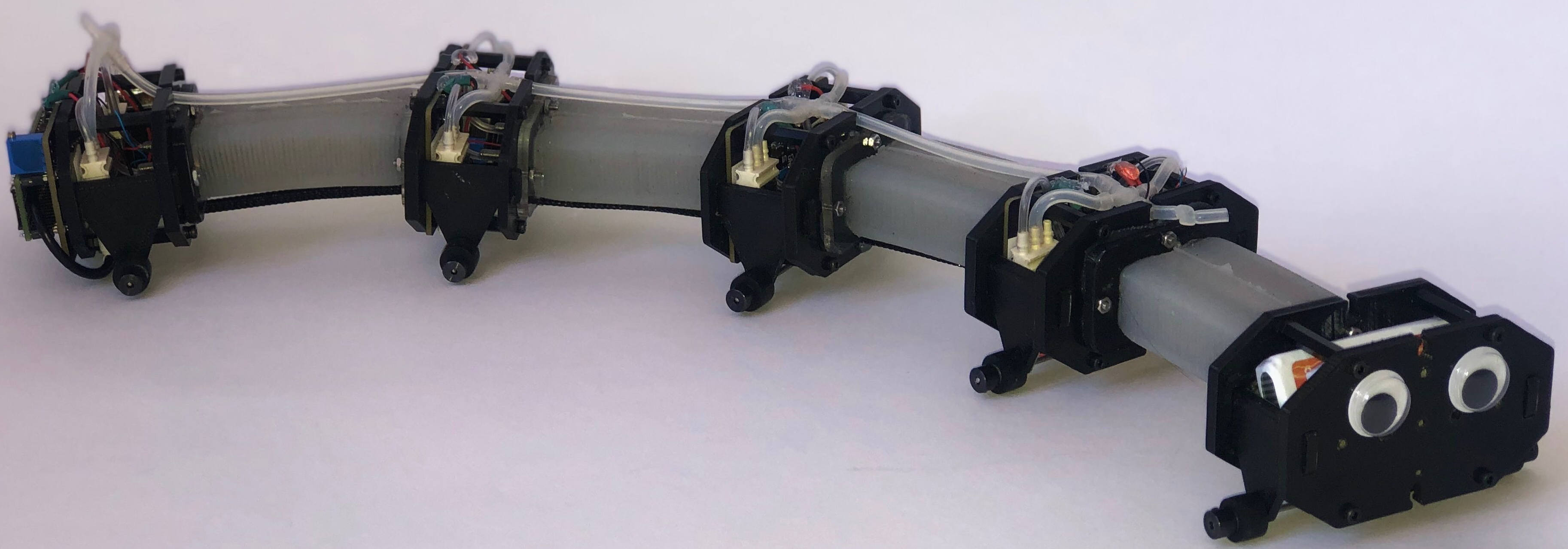}
            \small{(c)}
            \tcblower
            \centering
            \hspace{-2.3em}\includegraphics[width=1.2\linewidth]{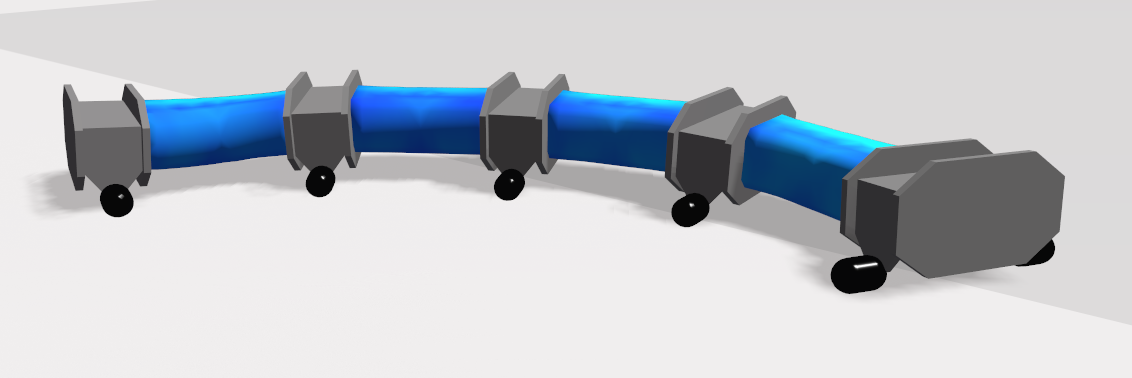}\\
            \hspace{-2.3em}\small{(d)}
            \end{tcolorbox}
    \end{tcolorbox}
    \caption{(a,b) Single soft link. The chambers are wrapped with a fiber reinforcement, preventing it from increasing in radius when pressure is applied. The center of the link contains an inextensible layer that prevents it from expanding in length \cite{snake_1}. The tiny spheres attached to the rigid plates are markers used for tracking the curvature in the experiments. (a) No pressure applied. (b) 8 psi applied on left chamber. (c) Full assembly of the robotic snake with four links. The main controller receives wireless commands from the computer, and passes it over to each slave controller, which activates the solenoid valves that release the pressure to the soft actuator. (d) Snake in simulation.}
    \label{fig:snake}
\end{figure}

The snake is made of soft bending actuation modules with integrated curvature sensing \cite{snake_1}, as shown in \figref{fig:snake}c. Each soft bending actuator segment is comprised of two soft linear actuators and an inextensible constraint layer between them. The soft links are made of silicone rubber with cavities wrapped in a fiber reinforcement, which limits the radial deformation when pressurized. In the center of the link, between the two chambers, there is a custom integrated curvature sensor, and a plastic film that inhibits linear extension. This set of constraints results in bending the entire soft module when one chamber is pressurized, as seen in \figref{fig:snake}b. The curvature sensor utilizes a magnet and a Hall effect sensor, mounted on a flexible circuit board \cite{snake_1}. Caps are attached to both ends of the actuator to seal the chambers and allow modular connections with other segments. The caps are made of two ABS plates sandwiching the rim of the silicone. The actuator is driven by two 3-2 (3-port, 2-state) binary solenoid valves, each connecting one pressure chamber to a common pressure source. 
The valves can either inflate or deflate a given actuator chamber. The pressure in each chamber is controlled using a 60 Hz Pulse-Width Modulation (PWM) on the valves, which are set to operate in complete antagonism, so when one chamber is inflating the other is always deflating. This reduces the number of required inputs per chamber to one, corresponding to the single (active) \ac{dof} of the bending actuator.

\section{Simulator}
\label{sec:simulator}

The continuous equations of motion for the multiphysics simulator are derived from Lagrangian mechanics, and are given in general form by the following, 

\begin{align*} 
\v{M}\qddot - \v{f}(\q, \qdot) - \v{J}_b^T\gv{\lambda}_b - \v{J}_n^T\gv{\lambda}_n - \v{J}^T_f\gv{\lambda}_f = \v{0}\\ 
\v{c}_b(\q, \v{p}) + \v{E}\gv{\lambda}_b = \v{0}\\
\v{0} \le \v{c}_n(\q) \perp \gv{\lambda}_n \ge \v{0}\\
\forall i \in \mathcal{A}, \quad \v{D}_i^T\qdot + \frac{|\v{D}_i^T\qdot|}{|\gv{\lambda}_{f,i}|}\gv{\lambda}_{f,i} = \v{0}\\
\forall i \in \mathcal{A}, \quad 0 \le |\v{D}_i^T\qdot| ~~ \perp ~~ \mu_i\lambda_{n,i} - |\gv{\lambda}_{f,i}| \ge 0 \\
\forall i \in \mathcal{I}, \quad \gv{\lambda}_{f,i} = \v{0}.
\end{align*}

These equations describe the motion of a generic dynamics system with frictional contact forces. 
The state of the system is described by a vector of generalized coordinates $\q \in \mathbb{R}^{n_d}$ with $n_d$ \ac{dof}s, determined by the number of particles and rigid bodies on the system. The inertial properties of the system are represented by the mass-matrix $\v{M} \in \mathbb{R}^{n_d\times n_d}$, with $\v{f}(\q, \qdot)$ a generalized force function that includes external and gyroscopic forces. The vector $\v{c}_b(\q)$ is a set of bilateral constraints of length  $n_b$, with $\gv{\lambda}_{b}$ the associated Lagrange multipliers. Elastic energy potentials are defined in terms of compliant constraints, here $\v{E} \in \v{R}^{n_b\times n_b}$ is a block-diagonal compliance, or inverse stiffness matrix as described by Servin et al. \cite{servin2006interactive}. The target pressures are grouped into the vector $\v{p}$, which are parameters to the actuation constraints described in section \ref{sec:simulator_constraints}. The contact and frictional forces are based on Coulomb's model, which defines an admissible cone of contact forces \cite{stewart1996implicit}. Here $\v{c}_n(\q)$ are unilateral contact constraints, with $n_c$ the number of contacts in the system, and $\lambda_{n,i}$ and $\mu_i$ the normal force Lagrange multiplier and friction coefficient for the $i$th contact respectively. The frictional forces for a contact are parameterized by $\gv{\lambda}_{f,i}$, with a corresponding basis $\v{D}_i$ that defines the surface tangent plane at the contact point. The active contact set is defined as $\mathcal{A} = \{i \in (1,\cdots, n_c) \mid \mu_i\lambda_{n,i} > 0\}$, with inactive contacts $\mathcal{I}$ being its complement. The constraint Jacobians $\v{J}_b, \v{J}_n$ contain the gradient of bilateral and normal constraint functions with respect to $\q$, and we define the set of frictional basis vectors as the matrix $\v{J}_f = [\v{D}_1, \cdots, \v{D}_{n_c}]^T$.


\subsection{Particles}
Each deformable link is modeled as a collection of particles connected by constraints. This is a flexible representation that allows fine-grained control over different sections of the soft body, while being efficient enough for real-time simulation. A particle with index $i$ adds three additional \ac{dof}s to the system,
\begin{align}
\q_i = \begin{bmatrix} x & y & z\end{bmatrix}^T.
\end{align}
Assuming a lumped mass model, each particle is assigned a fraction of the connected tetrahedral elements mass (Section \ref{sec:fem}). The mass-block for the particle is then given by $\v{M}_i = m\v{1}_{3}$, where $m$ is the particle mass, and $\v{1}_{3}$ is the $3$-dimensional identity matrix.

\begin{figure}[h]
    \centering
    \includegraphics[width=0.33\columnwidth]{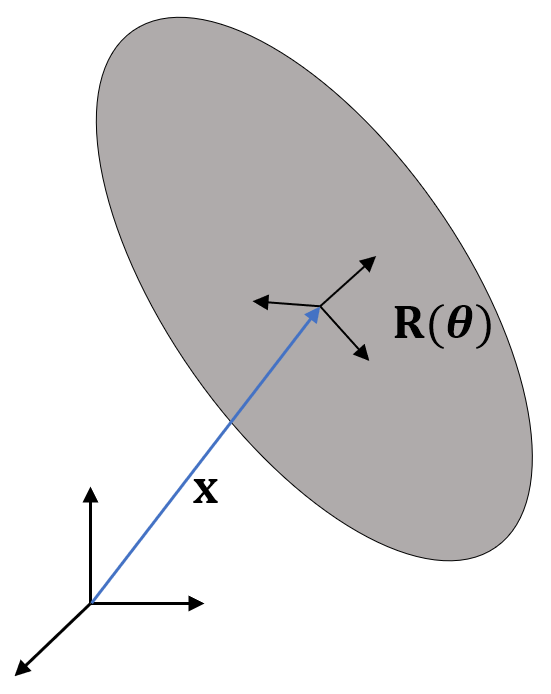}
    \caption{Rigid links and wheels are described by the translation of the body's center of mass from the origin $\v{x}$ and, it's orientation expressed as a quaternion $\gv{\theta}$.}
    \label{fig:rigid_frame}
\end{figure}

\subsection{Rigid Bodies}
We describe the state of a rigid body with index $i$ using a maximal coordinate representation consisting of the position of the body's center of mass, $\v{x}_i \in \mathbb{R}^3$, and its orientation expressed as a quaternion $\gv{\theta}_i = [\theta_1, \theta_2, \theta_3, \theta_4]^T \in \mathbb{R}^4$. We group these components together so the state sub-block for a single rigid body is
\begin{align}
\q_i = \begin{bmatrix}\v{x}_i^T & \gv{\theta}_i^T\end{bmatrix}^T.
\end{align}

\subsection{Constraints}
\label{sec:simulator_constraints}
\subsubsection{Actuation Constraints}
To perform actuation we constrain particles together through equations of the form,
\begin{align}
    c_{dist}(\q, p) = |\q_i - \q_j| - r\epsilon(p) = 0,
\end{align}
where $\q_i$ and $\q_j$ are particle positions, and $r$ is a rest length to maintain between them. The target pressure $p$ induces a strain $\epsilon(p)$ that adjusts the rest length and causes contraction or expansion. Assuming that deformation is linear with stress, and that it occurs primarily along the chamber's main axis, the amount of expansion/contraction is given by the following relation between material stiffness determined by the Young's Modulus ($Y$) and pressure $p$,

\begin{equation}
    \epsilon(p) = 1 + p/Y
    \label{eq:expansion}.
\end{equation}

Furthermore, we use distance constraints with constant rest length to model the structural stiffness in the deformable chamber, as described in Sec. \ref{sec:snakeconstraints}.\\

\subsubsection{Tetrahedral Finite-Elements}
\label{sec:fem}
In addition to distance constraints, tetrahedral finite-elements are used to model the solid chamber material. Assuming a constant strain element and a linear isotropic constitutive model, each tetrahedron defines a 6-dimensional constraint vector,

\begin{equation}
\v{c}_{tetra}(\q) + \v{E}_{tetra}\gv{\lambda} = \v{0},
\end{equation}
where $\v{c}_{tetra}(\q) = [\epsilon_{xx} ~ \epsilon_{yy} ~ \epsilon_{zz} ~ \epsilon_{yz} ~ \epsilon_{xz} ~ \epsilon_{xy}]^T$ is the vector of corotational strains in Voigt notation, and $\v{E}_{tetra}$ is the constant element compliance matrix, given by
\begin{align*} \v{E}_{tetra} = 
\frac{1}{V_e\,Y}\begin{bmatrix}
1 & -\nu & -\nu & 0 & 0 & 0 \\
-\nu & 1 & -\nu & 0 & 0 & 0 \\
-\nu & -\nu & 1 & 0 & 0 & 0 \\
0 & 0 & 0 & 1 + \nu & 0 & 0 \\
0 & 0 & 0 & 0 & 1 + \nu & 0 \\
0 & 0 & 0 & 0 & 0 & 1 + \nu\end{bmatrix}.
\end{align*}

where $V_e$ is the element volume, $Y$  and $\nu$ are the material Young's modulus and Poisson's ratio, respectively.\\

\subsubsection{Rigid Body to Particle Attachment}
In order to connect  soft links to  rigid bodies, an attachment constraint between a particle and a point on a rigid body is defined as follows,
\begin{equation}
    \v{c}_{attach}(\q) = \q_x + \v{R}(\q_\theta)\v{r} - \q_p = \v{0}.
    \label{eq:rigid_to_particle}
\end{equation}

This is a vector-valued function that adds three separate constraints, one for each $x,y,z$ axis respectively. Here $\q_x, \q_\theta$ are the rigid body position and orientation respectively, and $\q_p$ is the particle position. $\v{R}(\q_\theta)$ is the rotation matrix obtained from the body's orientation, and $\v{r}$ is the attachment point expressed in the body's local frame.

\subsubsection{Rigid Body Joints and Contact}

Along with the deformable sections, we model the articulated carriage as rigid bodies, with wheels connected to the main frame using hinge joints as described by \cite{shabana2009computational}. Contact between the wheels and the ground is modeled by non-interpenetration constraints of the form:
\begin{align}
c_n(\q) = \v{n}^T\left[\v{a}(\q) - \v{b}(\q)\right] \ge 0,
\end{align}
where $\v{n} \in \v{R}^3$ is the contact plane normal, $\v{a}$ and $\v{b}$ are points on a rigid or deformable body. Frictional forces are included using a Coulomb model derived from a principle of maximum dissipation that limits the contact forces to a cone. We refer  the readers to the survey paper by Stewart \cite{stewart2000rigid} for more detail.

\subsection{Time-Stepping}

The simulation is advanced in time with a first-order implicit time-discretization of the equations of motion similar to the method in \cite{todorov2010implicit}. An implicit discretization is chosen as it allows taking large time-steps and avoids constraint drift. At each time-step, the nonlinear system of equations resulting from the implicit discretization is solved using Newton's method. To solve the complementarity conditions associated with contact we use a non-smooth reformulation based on the Fischer-Burmeister function as described in \cite{munson2001semismooth}. Each Newton iteration requires the solution of a sparse-matrix equation of the form
\begin{align}\label{eq:schur_system}
\left[\v{J}\v{M}^{-1}\v{J}^T + \v{E}\right]\Delta\gv{\lambda} = \v{b}.
\end{align}
Where $\v{J} = [\v{J}_b^T ~ \v{J}_n^T ~ \v{J}_f^T]^T$ is the matrix of constraint Jacobians, $\v{E}$ is a block-diagonal compliance matrix that includes the tetrahedral compliance matrices, and $\v{b}$ includes the constraint function residuals evaluated at the current Newton iterate. This is a positive semi-definite system that we solve using the Preconditioned Conjugate Residual method (PCR) \cite{saad2003iterative}. This is an iterative Krylov method similar to Conjugate Gradient (CG) but with smooth error reduction, making it better suited for real-time applications with a fixed computational budget. Like CG, the primary computation cost of PCR is the performing sparse matrix-vector multiplications. However, these multiplications are highly parallelizable, and can be done efficiently by assembling $\v{J}, \v{M}, \v{E}, \v{b}$ on the GPU in compressed row-storage (CSR) format, and performing the multiplication with optimized kernels \cite{nvidia2014cusparse}. In our simulator we use a simple diagonal Jacobi preconditioner since it is trivial to parallelize.

\section{Soft robotic snake model}
\label{sec:snakeconstraints}

\begin{figure}
    \centering
    \begin{tcolorbox}[size = minimal,sidebyside, righthand width = .55\linewidth,sharp corners ,blanker, halign=center, halign lower=left]
    \centering
    \includegraphics[height=\columnwidth, angle=90]{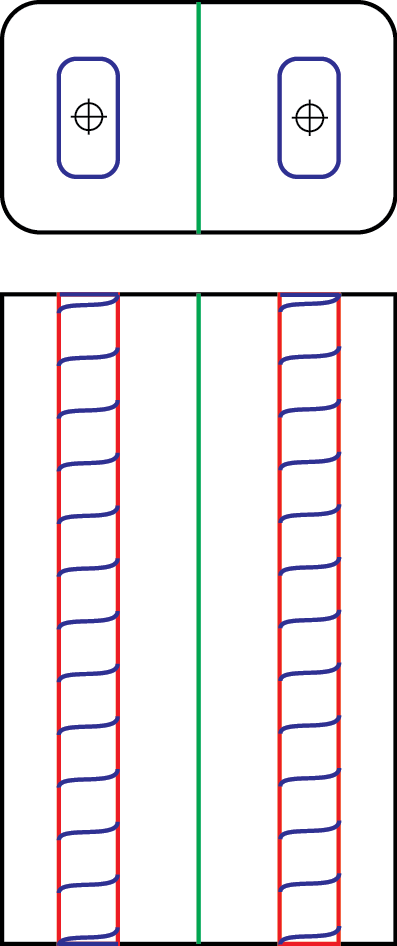}\\   
    \vspace{0.24cm}\small{(a)}
    \tcblower
    \begin{tcolorbox}[size = minimal, sidebyside, righthand width = .55\linewidth,sharp corners ,blanker, halign=center, halign lower=center]
    
    \centering
    \hspace{-2em}\includegraphics[width=1.5\columnwidth]{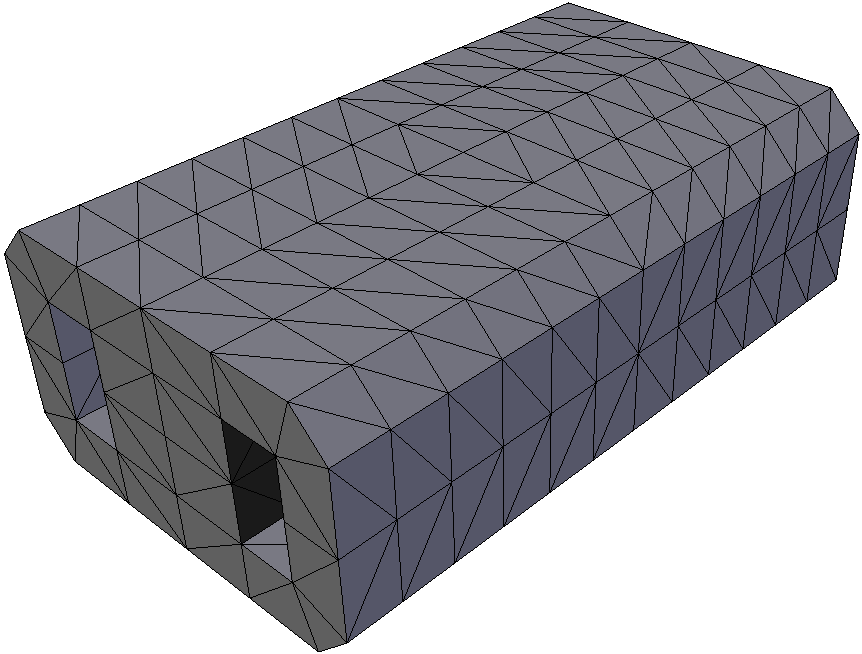}\\
    \vspace{0.23cm}\hspace{-2em}\small{(b)}
    
    \tcblower
    \centering
    \hspace{-2em}\includegraphics[width=1.2\columnwidth]{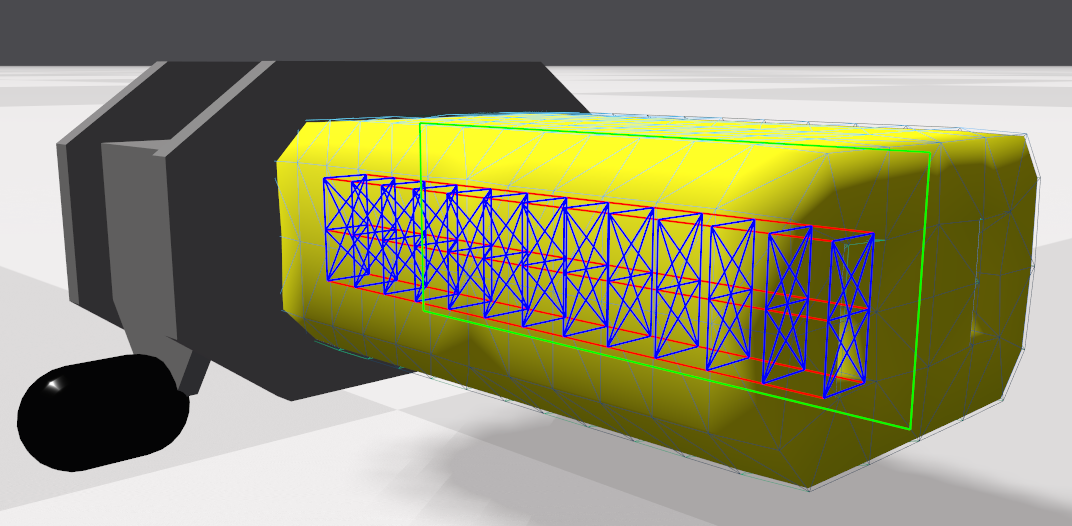}\\ \hspace{-2em}       \small{(c)}
    \end{tcolorbox}
    \end{tcolorbox}
    \caption{\small{(a)} Front and top view of chamber with constraints between particles on link. Green: stiff inter-particle constraint to limit chamber expansion. Blue: stiff constraints to ensure chamber radial perimeter is constant. Red: distance constraint used to expand chamber as overpressure is applied. \small{(b)} Soft link mesh. \small{(c)} Constraints displayed on simulation (best seen in digital format).}
    \label{fig:link_constraints}
\end{figure}

 The soft links of the snake robot are made of Ecoflex\texttrademark~00-30 silicone rubber which has material parameters $Y=66.243 \text{KPa}$, and $\nu=0.4999$ \cite{di2015stretch}. We construct a triangular mesh of the surface and tetrahedralize it using TetGen \cite{si2006quality}. The link mesh was created with evenly distributed particles, we do not explicitly represent the cavity with tetrahedral elements. The mesh was carefully constructed to provide a radially symmetrical tetrahedron structure, as seen in Fig. \ref{fig:link_constraints}. 
 

Since the inextensible layer in the center of the link has a deformation threshold that is  beyond the range of forces to be applied on the soft links, it is acceptable to model it as a non-deformable constraint between particles along the center plane. Similarly, the radial constraint on the chambers are defined as a set of inter-particle constraints over coplanar particles along the link. Although it would be possible to drive each link's expansion using surface pressure forces directly, the other constraints in the link allow us to simply control the chamber volume using constraints between particles along the primary axis of expansion.

Only one chamber on the link is active (\ie ~pressurized) at a time, so this set of actuation constraints only applies to the expanding chamber. \figref{fig:link_constraints} displays the constraints overlaid on the link. The link mesh was subdivided in 13 cross-sections along its length, in order to allow real-time computation, while maintaining good accuracy on the material deformation. 

The links are then connected to each other through the rigid bodies that contain the electronics necessary to control the snake robot. In addition, the rigid bodies are attached to the wheels via. hinge joints. The wheels provide contact with the floor and model the anisotropic friction that a real snake has from its scales.

\begin{table}[h]
    \centering
    \vspace{0.2cm}
    \begin{tabular}{l|c}
         \textbf{Type}& \textbf{Quantity}  \\
         \hline
         Rigid Bodies &  15 \\
         Particles & 1504 \\
         Distance Constraints  & 1460 \\
         Tetrahedral Finite Elements & 4536 \\
         Rigid Joints & 10 \\
         Particle Attachments & 217
    \end{tabular}
    \caption{size of the structure for one simulated snake}
    \label{tab:constraints_final}
\end{table} 
    
The type and number of all \acp{dof}, and constraints in the simulated snake are displayed in Table \ref{tab:constraints_final}. 
    
\subsection{Open-Loop Control}
The snake assembly consists of four links attached together, as seen in \figref{fig:snake}. An undulating motion that propels the snake is given by the control equation \ref{eq:snake_control}, which outputs the pressure for the link $i$.
\begin{equation}
    \label{eq:snake_control}
    a_i \equiv \min(1, \max(-1, (\sin(\omega t + \alpha i)+\phi))
    \, A.
\end{equation}

If $a_i$ is positive, the link will inflate one chamber of the link, while if $a_i$ is negative, it will inflate the other. The parameters $\omega$, $\alpha$, $\phi$, and $A$ are the base oscillation frequency, measured in $Hz$, the phase shift between links ($radians$), the offset value for the motion ($[-1,1]$, scalar), and the oscillation amplitude ($[0,p_{max}]$, Pa), respectively. The oscillation frequency dictates how fast the actuators will switch direction, and the phase delay between links is what generates the wave pattern that propels the snake forward. These parameters set the base undulating motion. By changing the offset $\phi$, applied to all links, the controller will inflate one side for longer than the other. This results in the snake propelling itself more to the opposite direction than the chambers that are more inflated, making the trajectory of the center of mass moves with a curvature radius determined by this offset and the friction with the ground. Finally, the amplitude $A$ limits the maximum pressure during the oscillation, thereby controlling the snake's speed. The parameters that make the snake move forward depend on the physical properties of the snake, such as weight, length, friction coefficient with the floor, and were determined experimentally in \cite{snake_1}. This controller is an open-loop method, which generates the forward motion and allows to make turns, but no feedback is given if the trajectory is deviating from the desired trajectory.

The pressure delivery is not instantaneous but limited by the maximum airflow allowed by the valves. Assuming the pressure source  can reliably maintain a constant output pressure $p_s$, the air flow $v$ to the chamber is given by \cite{white2008fluid},
\begin{equation}
    v^2 = \frac{2}{\rho}(p_t - p_s),
\end{equation}
where $\rho$ is the air density and $p_t$ is the pressure in the chamber. This means the pressure update is proportional to the square of the difference between the current pressure and the desired pressure. The pressure update for inflation in each step is then defined based on the difference $\Delta p_i$, in Eq. \ref{eq:delta_p_i}.
 \begin{align}
     \Delta p_i = \frac{a_i(t+h) - p_i(t)}{p_s},
     \label{eq:delta_p_i}
 \end{align}
 The deflation  releases pressure in the atmosphere while keeping its own pressure relatively constant due to the change of volume. Therefore, the deflation ratio should be close to linear up to a threshold $T_p$ when it is proportional to the over-pressure with a damping, 
\begin{align}
    p_i(t+h) = \begin{cases}
            p_i(t) + p_s{\Delta p_i}^2 k_{i} & \text{is inflating}\\
        p_i(t) - \min(p_i(t) k_{d},T_p) & \text{is deflating}
    \end{cases}    
\label{eq:deflation}
 \end{align}
 
  where 
  $k_{i}, k_d \in (0,1]$ are the inflation damping parameters, and are tuned according to the experimental data.


\section{Experimental Verification}
\label{sec:experiment}
\begin{figure*}[t]
\centering
\vspace{.2cm}
    \includegraphics[width=\linewidth]{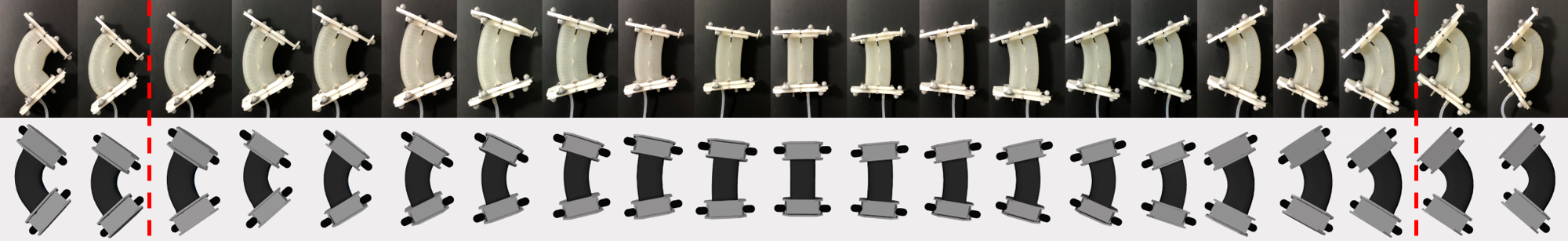}
    \caption{Visual comparison of link expansion after settling from -10 to 10 psi overpressure. Negative values mean the left chamber is inflated, while positive values are for the right chamber. The simulation displays high accuracy on the curvature up to 8 $psi$, where the dashed lines were traced. After that the pressure becomes excessive and the real link stops following the linear model (best seen in digital format). 
    }
    \label{fig:all_expansions}
\end{figure*}
\begin{table*}[t]
    \label{tab:benchmark}
	\centering
	\begin{tabular}{l|c|c|c|c|c|c|c|c|c|c|c} 
	Snakes (\#) & ${1}/{4}$ & 1 & 2 & 3 & 4 & 5 & 6 & 7 & 8 & 9 & 10 \\
	\hline
	\textbf{Assembly (ms)} &\small{ 0.74 }&	\small{1.11} &	\small{1.41} &	\small{2.33} &	\small{3.41} &	\small{5.59} &	\small{6.19} &	\small{6.45} &	\small{7.37} &	\small{8.48} &	\small{10.31} \\
	\textbf{Solve (ms)}  & \small{4.96} & \small{10.52} &\small{25.06 }& \small{31.90} & \small{43.12} & \small{50.41} & \small{58.63} &\small{ 66.95 }& \small{79.73} & \small{86.89 }& \small{95.57} \\
	\textbf{Total (ms)}  & \small {5.70} &\small { 11.63} & \small {26.47} & \small {34.23} & \small {46.53} & \small {56.00} & \small {64.82} & \small {73.4} & \small {87.10} & \small {95.37} &\small {105.88} \\
	\textbf{Total/Snake (ms)} & \small {-} & \small {\textbf{11.63}} & \small {13.23} &\small { 11.41} & \small {11.63} & \small {11.2} & \small {10.80} & \textbf{\small {10.48}} &\small { 10.88} & \small {10.59} & \small {10.58}
	\end{tabular}
	\caption{Benchmark results for our simulator}
	 \vspace{-.1cm}
\end{table*}

For all tracking experiments, the poses of the rigid links were captured using the motion capture system (MOCAP) by placing four markers on each rigid extremity so that it can collect their full poses. The MOCAP system contains 11 cameras surrounding the observable space of $4\times3~m^2$. 
This redundancy allows the information of links to be collected with high precision and minimal loss of tracking during the experiments. In order to eliminate remaining outliers, every experiment is repeated for 10 times unless otherwise mentioned. To ensure data sanity, every sample that deviates more than an expected maximum displacement from the collected values is pruned and not used in the analysis, as this type of samples results in a data collection with negligible standard deviation. \figref{fig:snake} shows the markers on the top corners of the rigid plates.

A friction coefficient of $\mu_i=1$ has been used for all experiments. We implement our simulator in CUDA and run it on a computer with an Intel Core-i7 8520K, 32GB of RAM, and one NVIDIA GTX 1080 Ti GPU. We use a fixed time-step of $h=0.0083s$, each time-step performs 4 Newton iterations, with each linear system solved approximately using 20 PCR iterations to ensure a fixed computational cost.


\subsection{Quasi-static Verification}

The first experiment on the simulation is to verify whether the pressure actuator follows the same geometrical behavior as the real link. For this experiment, the curvatures of the real links were obtained by subtracting the yaw of the rigid connectors attached to each soft link, for the varying over-pressures (the pressure that exceeds the resting atmospheric pressure) from $0$ to $10$ psi, moving up with steps of $1$ psi for both directions on the link, and averaged  over $300$ samples. Negative values show the inflation of left chamber and positive values are for the right chamber.  

\begin{figure}[H]
    \centering
    \includegraphics[width=1.0\columnwidth]{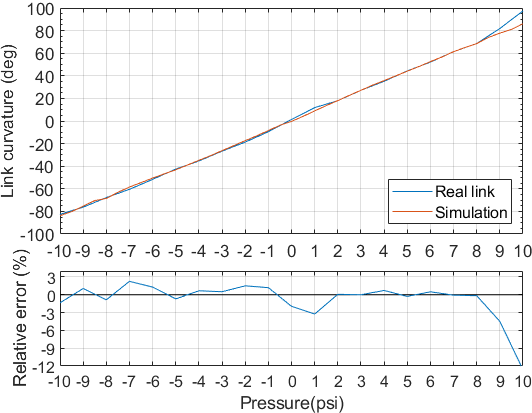}
    \caption{Curvature vs. Pressure for simulation and a real link. The top plot shows that the curvature follows a linear relationship with the pressure applied. The bottom plot is the relative error.}
    \label{fig:plot_curvature}
\end{figure}


From \figref{fig:plot_curvature}, it can be seen that the curvature increases linearly with the pressure within a range. Particularly, the spring model is able to closely match the real curvature, and accurately follows the linear model up to 8psi. Ecoflex 00-30 Young modulus is 66 kPa, at 9 psi (62 kPa) it nears 2x expansion, which is the limit at which the material is linear. This behavior is also  clearly observed by the relative error plot. 
When the pressure exceeds 8 psi, the real link starts to bulge over imperfections in the manufacturing, and on the opposite side, it folds in itself, resulting in a deviation from the linear model. Besides, since the extra pressure is forcing over the imperfections of the manufacturing process, there is a potential risk of damaging the links in the long term. For these reasons, it was deemed that the safe pressure threshold shall be 8 psi, and all the remaining tests were restricted to up to that range.  


\subsection{Dynamic Verification}
\begin{figure}
    \centering
    \includegraphics[width=1.0\columnwidth]{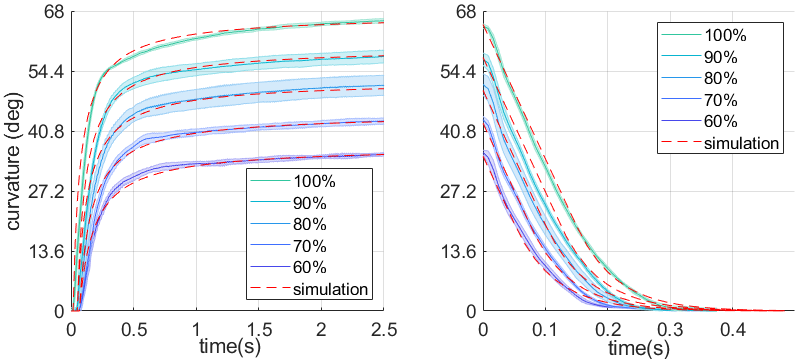}
    \caption{Step analysis for chamber inflation (left) and deflation (right) with approximate response from simulator.}
    \label{fig:plot_step}
    \vspace{-0.2cm}
\end{figure}

The dynamic verification starts with the analysis to the step response on the actuators. A single link was used to capture the rate at which it inflates and deflates from rest to $60-100\%$ at $10\%$ steps. These trials were used to tune the gains $k_{i} = k_{d} = 0.23$ and $T_p = 0.68$. The results are seen in \figref{fig:plot_step}.

\begin{figure}[h]
    \centering
    \includegraphics[width=\columnwidth,]{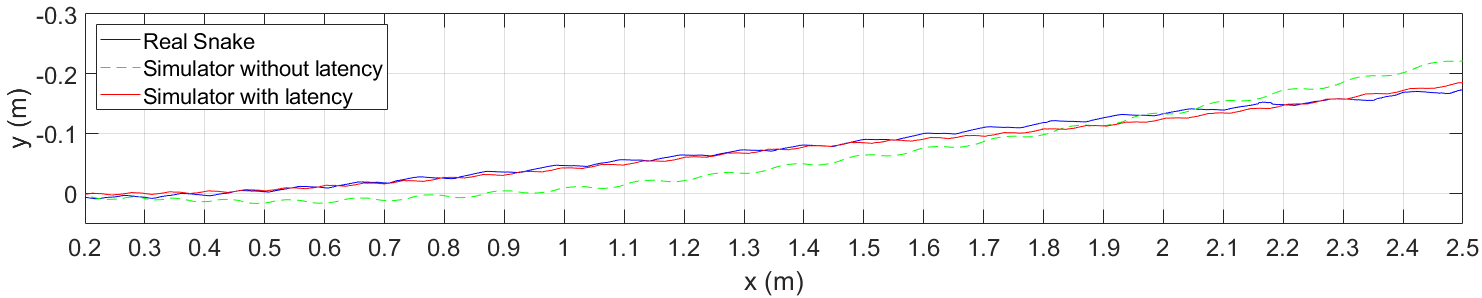}
    \caption{Trajectory comparison between collected data (blue), simulator without actuator latency (green), and with latency (red). The latency model in the pressure update makes the simulator's trajectory be closer to the real snake under same conditions (best seen in digital format).}
    \label{fig:plot_trajactory}
    \vspace{-0.4cm}
\end{figure}

The simulator trajectory is then tested with an open-loop generator from Eq. \ref{eq:snake_control}  and compared with the real snake robot using the following parameters $\omega = 2Hz$, $\alpha = \pi/2$, $\psi = 0$ and $A = p_{max} = 8psi$.  Since it's an open-loop control, it is expected that due to model inaccuracies and unmodeled dynamics the simulator will have error accumulated along the trajectory. As a result, the simulated trajectory may diverge from the actual trajectory over time. The comparison was made using the center of mass of the snake.  The divergence can be addressed with the use of closed-loop feedback control for both simulated snake and the real snake robots.
However, from \figref{fig:plot_trajactory}, it can be seen that starting from the same initial condition, the simulated trajectory closely matches the real trajectory  for the execution time. 
The inclusion of the latency model for the pressure update makes the trajectory of the center of mass go in the same overall direction and amplitude as the real snake. 

\subsection{Benchmarking}

In order to test scalability of the system, we benchmark simulation times for a single soft link, a full snake with 4 links, and up to 10 snakes, each with 4 links. The linear system for a single Newton iteration corresponding to a single snake is a sparse matrix of roughly $30000\times30000$. Each frame is sub-divided in two sub-steps for the constraint solver, and the optimization runs with 20 iterations each. The total per-frame simulation times are in Table \ref{tab:benchmark}. From the results, it can be seen that simulation time increases linearly with the number of links and snakes. Linear scaling is expected, since a single link saturates the GPU. On average, it takes less than $12\text{ ms}$ to simulate each snake, with optimal performance obtained when simulating seven snakes together. One single snake takes $8.9 s$ per frame running at 3.3GHz on an Intel Core i7 5820k.

 

 \section{Conclusion}

We have presented a dynamical model for simulating 1D pneumatic actuators and a framework to simulate it in a multi-physics environment in real time. By comparing the simulation with a real soft robotic snake, it is demonstrated that the simulated snake produces real-time high-fidelity results even in complex scenarios involving a mixture of hybrid soft bodies, rigid bodies, and friction contacts. In the open loop control analysis, it becomes clear that the simulator doesn't perfectly model the real dynamics; however, it remains in close performance to the real snake. Our next step is now apply a control trained and tested in simulation on the real snake.
The transfer from simulation to reality, while still maintaining stability, can also be facilitated by domain randomization techniques \cite{Sim-to-real}. We demonstrated the use of GPU in accelerating high-fidelity simulation for soft robots and present a framework that can generalize to other soft robotic systems. Our future research is to develop learning-based control to train fast and stable snake gaits in a range of terrains, as well as obstacle-aided navigation, and learning specific motion primitives from demonstration \cite{ijspeert2008central, ijspeert2013dynamical}. Another step is also study the model different types of soft actuators such as PneuNet \cite{trivedi2008soft}, and plan to publish data for our snake model to enable an open platform for experimentation and improvement.

\bibliographystyle{IEEEtran}
\bibliography{IEEEabrv,ms}

\end{document}